\documentclass[journal]{IEEEtran}
\usepackage{xcolor,soul,framed}
\colorlet{shadecolor}{yellow}
\usepackage[pdftex]{graphicx}
\graphicspath{{../pdf/}{../png/}{../jpg/}}
\DeclareGraphicsExtensions{.pdf,.png,.jpg}
\usepackage[cmex10]{amsmath}
\usepackage{array}
\usepackage{mdwmath}
\usepackage{mdwtab}
\usepackage{eqparbox}
\usepackage{url}
\usepackage{flushend}
\usepackage{tikz}
\usetikzlibrary{plotmarks}
\usetikzlibrary{arrows}
\usetikzlibrary{arrows.meta}
\usetikzlibrary{patterns}
\usepackage{pgfplots}
\usepgfplotslibrary{patchplots}
\pgfplotsset{compat=newest}
\pgfplotsset{plot coordinates/math parser=false}
\pgfplotsset{tick label style={font=\footnotesize},
						 label style={font=\footnotesize},
						 legend style={font=\footnotesize}}
\usepackage{grffile}
\newlength\figureheight
\newlength\figurewidth
\begin{document}
\bstctlcite{IEEEexample:BSTcontrol}

\title{Jacobian Methods for Dynamic Polarization Control in Optical
Applications}

\author{Dawei Wang, Kaiqin Lai, Ying Yu, Qi Sui, and Zhaohui Li%
\thanks{This work was supported in part by the National Key Research and
Development Program of China (2019YFB1803502), in part by the National
Natural Science Foundation of China (U2001601), and in part by the
Key-Area Research and Development Program of Guangdong Province
(2018B010114002, 2020B0303040001). (\emph{Corresponding
author: Dawei Wang, Kaiqin Lai})}%
\thanks{D. Wang and Z. Li are with Guangdong Provincial Key Laboratory of
Optoelectronic Information Processing Chips and Systems, School of
Electronics and Information Technology, Sun Yat-sen University,
Guangzhou 510006, China, and also with Southern Marine Science and
Engineering Guangdong Laboratory (Zhuhai), Zhuhai 519000, China
(e-mails: wangdw9@mail.sysu.edu.cn; lzhh88@mail.sysu.edu.cn).}
\thanks{K. Lai and Y. Yu are with Guangdong Provincial Key Laboratory of
Optoelectronic Information Processing Chips and Systems, School of
Electronics and Information Technology, Sun Yat-sen University,
Guangzhou 510006, China (e-mails: laikq5@mail2.sysu.edu.cn).}%
\thanks{Q. Sui is with Southern Marine Science and Engineering Guangdong
Laboratory (Zhuhai), Zhuhai 519000, China (e-mails:
suiqi@sml-zhuhai.cn).} }

\markboth{Manuscript Submitted to Journal of Lightwave Technology on
\today} {}

\maketitle

\begin{abstract} Dynamic polarization control (DPC) is beneficial for
many optical applications. It uses adjustable waveplates to perform
automatic polarization tracking and manipulation. Efficient algorithms
are essential to realizing an endless polarization control process at
high speed. However, the standard gradient-based algorithm is not well
analyzed. Here we model the DPC with a Jacobian-based control theory
framework that finds a lot in common with robot kinematics. We then give
a detailed analysis of the condition of the Stokes vector gradient as a
Jacobian matrix. We identify the multi-stage DPC as a redundant system
enabling control algorithms with null-space operations. An efficient,
reset-free algorithm can be found. We anticipate more customized DPC
algorithms to follow the same framework in various optical systems.
\end{abstract}



\section{Introduction}
\IEEEPARstart{P}{olarization} management is an important topic in
multiple areas of optics, such as optical sensing \cite{caucheteur2017},
optical communication \cite{martinelli2006}, and quantum information
\cite{ma2012}. One extensively researched type of the polarization
management is called dynamic polarization control (DPC), which played a
crucial role in the early development of coherent optical fiber
communication \cite{noe1988} and has been successfully utilized in
recent studies of self-homodyne coherent optical detection
\cite{gui2021real} and quantum key distribution \cite{jouguet2013exp}.
The DPC typically performs transformation from an arbitrary input state
of polarization (SOP) to a desired output SOP via in-line control, a
useful feature to perform tasks such as polarization alignment,
polarization demultiplexing, and polarization stabilization in various
optical systems.

\begin{figure}[t]
\centering
\includegraphics[width=.95\columnwidth]{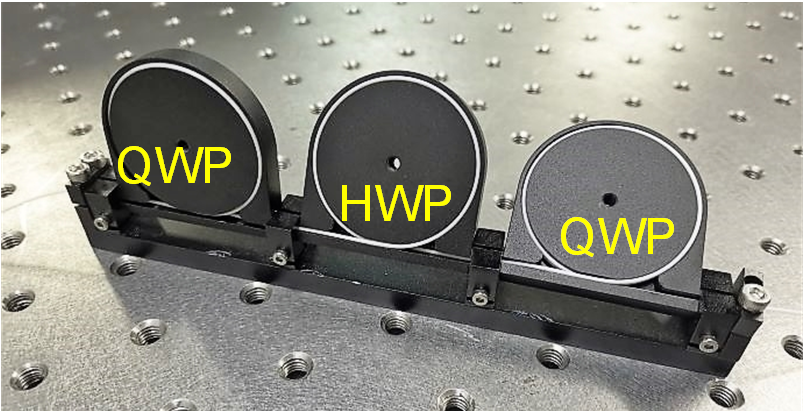}
\caption{Manual polarization controller with optical fiber coils working
as rotatable waveplates. QWP: quarter-wave plate. HWP: half-wave plate.}
\label{f1}
\end{figure}

Numerous DPC schemes have been proposed and tested, almost all of which
can be modeled as a system of waveplates. Fig. \ref{f1} shows the
picture of a widely used manual fiber polarization controller, which is
equivalent to three rotatable waveplates connected in series. Most DPC
schemes use similar structure but have different physical realizations
of the waveplates and also requrie additional control circuits. Examples
of such schemes include the one based on fiber squeezer
\cite{aarts1989}, liquid crystal, lithium niobite (LiNbO$_3$) crystal
\cite{noe2016}, metasurfaces \cite{park2017}, and plasma
\cite{turnbull2016}. Although seemingly different, they can be
effectively modeled as waveplates with variable thickness and/or
adjustable principal axes \cite{hirabayashi2003,van1993}. Notably, the
rapidly developing photonics integration provides both ideas and
techniques for realizing smaller and faster polarization controllers.
Several platforms on which integrated polarization control devices with
impressive performance have been reported
\cite{sarmiento2015,wang2021mach,lin2022,liu2022high}.

The DPC tracking speed is of primary concern among several parameters.
It relies on the optical components with fast response to external
control electronics and control algorithms with low latency. Fast
polarization response (on the order of $10^5$ rad/s) can be realized
via, e.g., the Pockels effect of LiNbO$_3$ crystals. A well-designed DPC
algorithm should be able to track the input SOP, which may vary randomly
and indefinitely, with control signals of a limited range. Control
signal reset near the range limit is nonetheless allowed, provided that
the output SOP does not change during the reset process. The most widely
used gradient descent algorithm uses a dithering technique to estimate
the current SOP gradient (with respect to the control signal) with
sophisticated signal-reset strategies \cite{noe1988}. The dithering is
simple to implement but would inevitably suspend the control loop
repeatedly. The signal reset is particular to the type of DPC component
and adds considerable complexity to programming. Other control
algorithms also exist, such as greedy linear descent \cite{ma2020auto},
genetic algorithm \cite{mamdoohi2012} and particle swarm optimization
algorithm \cite{chen2013}, but they are considerably more complex than
the gradient-based one.

The DPC is a typical system that fits the Jacobian-based control theory
framework, known as the Resolved Motion Rate Control in robotics
\cite{whitney1969}\cite[chap. 8]{corke2017}. When the control targets
(such as the position of controlled object) are viewed as a
vector-valued function of several free variables (force, voltage,
temperature, etc.), the Jacobian is the matrix of all its first-order
partial derivatives which is useful for linearizing the control problem.
In the DPC problem, the target Stokes vector as a function of the free
parameters of a multi-stage DPC device can also be linearized via the
Jacobian. However, there still lacks detailed analyses of the Jacobian
condition for a general DPC model in the literature. In contrast, this
is a well-studied topic in robotics \cite{siciliano1990}. We find that
many control techniques that have been widely used in robotics can be
analogously designed and applied to the DPC problem despite distinct
application scenarios and requirements. In this article, we give a
detailed analysis of the Jacobian condition based on a generalized DPC
model. We further introduce and study several standard Jacobian-based
control methods for the polarization control applications. We
successfully prove that the Jacobian of a general multi-stage DPC device
is almost always rank deficient. We also find that the gradient
projection method enables reset-free endless polarization control by
taking advantage of the null space of an underdetermined DPC Jacobian.
In view of its generality, we anticipate more variants of DPC algorithms
with distinct features in the same theoretical framework.

\section{The DPC problem} In the classical regime, a fully polarized
light is in a definite polarization state that can be described by a
real-valued Stokes vector $S=[s_1,s_2,s_3]^T$ with $T$ for transpose and
the property $s_1^2+s_2^2+s_3^2=1$. Therefore, all possible polarization
states can be represented uniquely as a point on a sphere, namely the
Poincar\'{e} sphere. The transformation between the input states
$S_\text{in}$ and the output state $S_\text{out}$ of a polarization
control element (i.e., a general waveplate) is then equivalent to the
3-dimensional rotation, $S_\text{out} = MS_\text{in}$, where the
rotation matrix $M$, known as the Mueller matrix, is an orthogonal
matrix if the polarization control is lossless. As examples, when the
three orthogonal axial vectors $S_1=[1,0,0]^T$, $S_2=[0,1,0]^T$, and
$S_3=[0,0,1]^T$ form a right-handed coordinate system, the waveplates
that rotate $S_\text{in}$ by an angle $\theta$ about $S_1$, $S_2$ and
$S_3$ are described respectively by the elemental rotation matrices
\begin{align*}
R_1 = \begin{bmatrix}
1 &0 &0\\ 
0 &\cos(\theta) &-\sin(\theta)\\
0 &\sin(\theta) &\cos(\theta) \end{bmatrix}
R_2 = \begin{bmatrix}
\cos(\theta) &0 &\sin(\theta)\\ 
0 &1 &0\\
-\sin(\theta) &0 &\cos(\theta)\end{bmatrix}
\end{align*}
and
\begin{equation*}
R_3 = \begin{bmatrix}
\cos(\theta) &-\sin(\theta) &0\\ 
\sin(\theta) &\cos(\theta) &0\\
0 &0 &1\end{bmatrix}\,.
\end{equation*}
The rotation direction follows the right-hand rule such that the axial
vector coincides with the angular velocity vector. A general
representation of 3d rotation matrix is given by
\begin{equation}
M = I + [\sin(\theta)](\vec{r}\times) 
+ [1-\cos(\theta)](\vec{r}\times)^2
\end{equation}
where $\vec{r}$ is the unit axial vector and $\vec{r}\times$ is the
cross product matrix of $\vec{r}$. Note that all the entries of matrix
$M$ are real-valued.

To convert an arbitrary $S_\text{in}$ into a designated $S_\text{out}$
means the polarization controller needs to be reconfigurable for
realizing all possible Mueller matrices. In practice, this is achieved
via composition, i.e., a system consists of multiple serial-linked
polarization control components having different characteristic rotation
matrices. As a result, the $m$-stage DPC is modeled as a sequence of
non-commutative polarization rotations, i.e., $S_\text{out} = M_m
M_{m-1} \cdots M_1 S_\text{in}$, which represents a trajectory
connecting the input and output SOP points on the Poincar\'{e} sphere. A
common rule for the composition is \emph{Euler angles}. E.g., an
arbitrary 3d rotation can be written as a combination of three elemental
rotations $M = R_1(\theta_3) R_3(\theta_2) R_1(\theta_1)$ where
$\theta_1$, $\theta_2$, and $\theta_3$ are three angles to be determined
by the control logic. The reverse process of finding the three angles
suffers from nonuniqueness and singularities in the motion of a rigid
body, which uses intrinsic rotations such as the roll-pitch-yaw angles.
However, the DPC stages have independent axes and hence deal with
\emph{extrinsic rotations} in the world coordinate frame. I.e., one
should picture the DPC process as a series of rotations of the
Poincar\'{e} sphere.

Note that although each $M$ can have multiple parameters that we can
control, the Euler-angles composition uses one control variable for each
DPC stage, which is very common in practical design due to easy
implementation. The DPC of this type needs at least 3 stages (3 degrees
of freedom (DOF)) to specify an arbitrary 3d rotation. But the SOP
points on the Poincar\'{e} sphere have only two DOFs, i.e. the longitude
and latitude, or the azimuth and ellipticity\footnote{Imagine the Stokes
vector as an airplane located at the origin of Poincar\'{e} sphere
pointing outwards. the DPC is similar to the aircraft attitude control
but simpler because the Stokes vector has no meaningful \emph{roll}
motion.}. We say that the DPC with three or more stages is inherently
redundant because it can realize infinitely many rotations (one is
\emph{geodesic}) that connect a given pair of SOP points on the
Poincar\'{e} sphere.


\begin{figure}[t]
\centering
\includegraphics[width=.95\columnwidth]{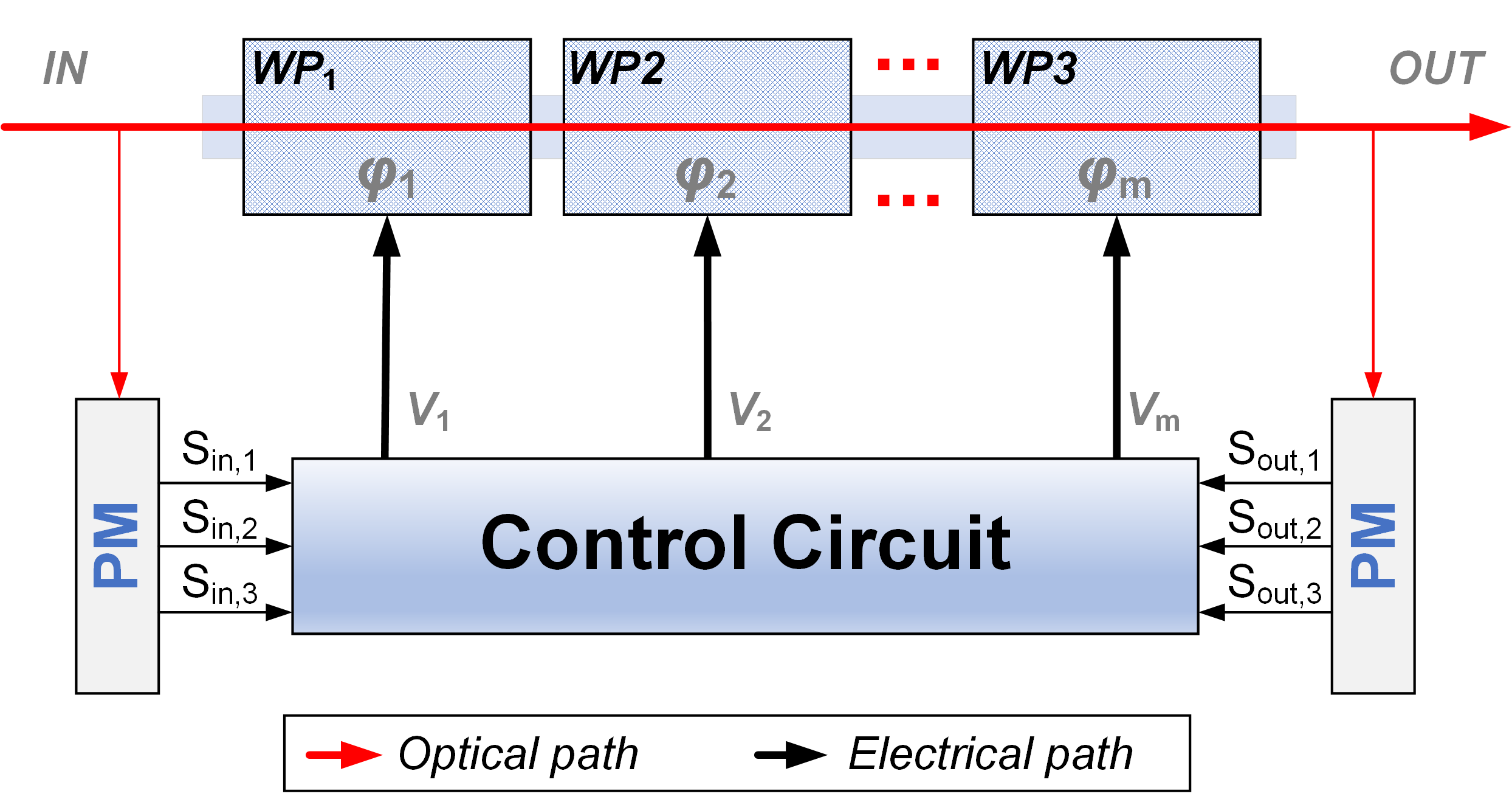}
\caption{Block diagram of the dynamic polarization control system with
serial-linked waveplates and Jacobian-based control loops. The
waveplates have key parameters $\varphi_{1,2,...}$ for altering the SOP
at the output. $V_{1,2,...}$ are the control voltages produced by the
control circuit. Small portions of light are decoupled from both the
input and the output for Stokes parameters measurement. WP: waveplate.
PM: polarimeter.}
\label{f2}
\end{figure}

To formalize the DPC problem, consider an $m$-stage DPC with
voltage-driven constituent waveplates modeled as Mueller matrices
$M_1,...,M_m$, which are parameterized with scalar external control
signals $\varphi_1,...,\varphi_m$, respectively. Define the vector
$\boldsymbol{\varphi} \equiv [\varphi_1,...,\varphi_m]^T$. The critical
step is to model the DPC process as the transformation from the
$m$-dimensional $\boldsymbol{\varphi}$-space to the $3$-dimensional
Stokes space. Namely,
\begin{equation}
S_\text{out} = \underbrace{M_m(\varphi_m) M_{m-1}(\varphi_{m-1})
\cdots M_1(\varphi_1) S_\text{in}}_{f(\boldsymbol{\varphi})} \,,
\label{eq:ctrl_mod}
\end{equation}
where the Stokes vector is arguably the best choice to represent the
2-DOF control task for practical reasons of easy measurement. Other
representation such as the longitude and latitude may not be well
defined at certain locations in the task space (e.g., two poles of the
Poincar\'{e} sphere).

In general, the mapping $f:\boldsymbol{\varphi} \to S_\text{out}$ is
nonlinear and determined by definition by both the input SOP and the
constituent Mueller matrix. The DPC problem seeks a solution by
reconfiguring all the waveplates via adjusting $\boldsymbol{\varphi}$ to
reach the target $S_\text{out}$ with varying $S_\text{in}$, subject to
the condition that each of $\varphi_i,\,i=1,...,m$, is limited to a
finite range. It is often the case that ${\varphi}_i$ is linearly
related to the rotation angle about a fixed axis in the $i$th stage. We
can also realize adjustable axis by combining several fixed-axis stages.
E.g., the combination $R_3(\phi)R_1(\theta)R^T_3(\phi)$ represents the
rotation of angle $\theta$ around an $\phi$-oriented axis in the $S_1$-$S_2$
plane. Care must be taken to obtain an accurate model especially when
the voltages are nonlinearly related to the adjustable axes.

The DPC problem formulated above closely resembles the arm-type robot
kinematics, where the $S_\text{out}$ can be considered as the state of
the robot end-effector, and the $m$ waveplates are the $m$ serial-linked
robot arms \cite{corke2017}. Both waveplates and robot arms have
physical limitations in their maximal tuning ranges. Given $S_\text{in}$
and $\boldsymbol{\varphi}$, the output state $S_\text{out}$ can be
computed via (\ref{eq:ctrl_mod}), which is termed \emph{forward}
kinematics in robotics (the mapping from robot configuration to
end-effector pose, see \cite[sec. 7.1]{corke2017}). Finding the required
control signals for realizing a specific $S_\text{out}$ corresponds to
the \emph{inverse} kinematics problem in robotics (the inverse mapping
of forward kinematics, see \cite[sec. 7.2]{corke2017}). The two problems
(DPC vs. robotics) differ in the dimensionality of their task spaces
(2 DOF in DPC vs. 6 DOF in most robotics).

It is in general difficult to solve for $\boldsymbol{\varphi}$ by direct
inversion of the mapping $f$. Also, the redundant DPC means we may find
infinite solutions for a given SOP target. However, the model can be
\emph{linearized} when small variations $\Delta\boldsymbol{\varphi}$ in
the control signals are considered, in which cases the model
(\ref{eq:ctrl_mod}) becomes
\begin{equation}
J \Delta\boldsymbol{\varphi}
= \Delta S_\text{out}\,,
\label{eq:ctrl_mod_lin}
\end{equation}
where $J$ is the $3 \times m$ Jacobian matrix defined as $J = \partial f
/\partial \boldsymbol{\varphi}$ evaluated at the point of linearization,
i.e.,
\begin{equation}
J = \begin{bmatrix}
\frac{\partial s_{1,\text{out}}}{\partial \varphi_1} &
\frac{\partial s_{1,\text{out}}}{\partial \varphi_2} &
\cdots &
\frac{\partial s_{1,\text{out}}}{\partial \varphi_m} \\
\frac{\partial s_{2,\text{out}}}{\partial \varphi_1} &
\frac{\partial s_{2,\text{out}}}{\partial \varphi_2} &
\cdots &
\frac{\partial s_{2,\text{out}}}{\partial \varphi_m} \\
\frac{\partial s_{3,\text{out}}}{\partial \varphi_1} &
\frac{\partial s_{3,\text{out}}}{\partial \varphi_2} &
\cdots &
\frac{\partial s_{3,\text{out}}}{\partial \varphi_m}
\end{bmatrix},
\end{equation}
$\Delta\boldsymbol{\varphi}$ and $\Delta S_\text{out}$ are implicitly
the rates of coordinate change in $\boldsymbol{\varphi}$-space and
Stokes space respectively, as they are variations over a small amount of
time (consider dividing both sides of (\ref{eq:ctrl_mod_lin}) by $\Delta t$).

To form a closed-loop control, $\Delta S_\text{out}$ is taken to be the
error between the current and the target SOP, i.e.,
\begin{equation}
\Delta S_\text{out}[k]
= S^{*}_\text{out} - S_\text{out}[k]\,,
\label{eq:ctrl_mod_err}
\end{equation}
where $S_\text{out}[k]$ is the measured Stokes vector at time instance
$k$. Then, the optimal $\Delta\boldsymbol{\varphi}$ can be found based
on the linear model (\ref{eq:ctrl_mod_lin}) with methods we shall
introduce in the next section. The waveplates are driven by the updated
signal
\begin{equation}
\boldsymbol{\varphi}[k+1] = \boldsymbol{\varphi}[k]
+ \Delta\boldsymbol{\varphi}[k]\,.
\label{eq:ctrl_mod_upda}
\end{equation}

The schematic of the Jacobian-based DPC is depicted in Fig. \ref{f2}.
Light is partially decoupled from both the input and the output optical
paths to measure the current $S_\text{in}$ and $S_\text{out}$. A
polarization measuring device based on photonics integration can be
found in \cite{lin2022}. This is however optional at the output as it is
possible to directly compute $S_\text{out}$ from (\ref{eq:ctrl_mod}).
The control circuit computes the Jacobian, solves the control model
(\ref{eq:ctrl_mod_lin}) to obtain the best $\Delta\boldsymbol{\varphi}$,
and drives the waveplates with updated voltages. Since knowing the
waveplate matrices and $S_\text{in}$ determines the Jacobian completely,
no dithering is needed if the closed-form Jacobian can be obtained.

\section{Jacobian methods}
\subsection{Direct or pseudo inversion} The condition of the Jacobian
matrix is critical for solving the linear model (\ref{eq:ctrl_mod_lin}).
A $3$-stage DPC has a $3\times 3$ square Jacobian matrix. If the matrix
is invertible, the control signals variation can be found easily via
$\Delta\boldsymbol{\varphi} = J^{-1}\Delta S_\text{out}$, where
$(\cdot)^{-1}$ stands for matrix inverse and $\Delta S_\text{out}$ is
given by (\ref{eq:ctrl_mod_err}). Consider $\boldsymbol{\varphi}$ as
voltages that produce linearly-related rotation angles only. It turns
out that all 3-stage DPC modeled by (\ref{eq:ctrl_mod}) has
noninvertible Jacobian of the form
\begin{align}
J 
&= \begin{bmatrix} M_3M_2(\vec{r}_1\times S') & M_3(\vec{r}_2\times S'') 
& {\vec{r}_3\times S'''}
\end{bmatrix}\,,
\label{eq:jacob}
\end{align}
where $\vec{r}_{1,2,3}$ are the rotation axes of Mueller matrices
$M_{1,2,3}$, respectively, whereas $S'$, $S''$, and $S'''= S_\text{out}$
are the Stokes vectors out of the first, the second, and the third stage
of DPC, respectively. The rank of Jacobian is $\le 2$ regardless of the
input SOP or the DPC configurations since all the column vectors of $J$
are perpendicular to $S_\text{out}$. The singularity is explicit
whenever $S'$, $S''$, or $S'''$ coincides with the rotation axis of the
corresponding stage, which has consequences in the DPC performance (see
next section). The results hold when ${\varphi}_i$ is an arbitrary
(linear or nonlinear) function of the corresponding rotation angle.
Moreover, by using the concept of polarization dynamic eigenstates
\cite{shieh2001dynamic}, the results can be generalized to the cases
where $\boldsymbol{\varphi}$ is related to any type of control
parameters that causes SOP changes at the DPC output. The form of (\ref{eq:jacob})
is a general result since the dynamic eigenstate $\vec{r}$ of matrix
$M$ follows the rule \cite[equ. 14]{shieh2001dynamic}
\begin{align}
\frac{dM}{d\varphi} = \vec{r}\times M,
\label{eq:des}
\end{align}
where $\varphi$ is the key parameter causing the SOP change.

To get around the singularity problem, we can use alternative methods to
find $\Delta\boldsymbol{\varphi}$, such as the transposed Jacobian,
$\Delta\boldsymbol{\varphi} = J^{T} \Delta S_\text{out}$ with
$(\cdot)^T$ for transpose, the damped Jacobian,
$\Delta\boldsymbol{\varphi} = {(J + \lambda I)}^{-1} \Delta
S_\text{out}$ where $\lambda$ is a proper constant and $I$ the identity
matrix, or the pseudo inverse Jacobian $\Delta\boldsymbol{\varphi} =
J^{+} \Delta S_\text{out}$ with $(\cdot)^+$ for the Moore-Penrose
inverse. The pseudo inverse solution has the feature of being minimum
norm among all viable solutions.

The square $J$ being singular means the linear model
(\ref{eq:ctrl_mod_lin}) is underdetermined (or \emph{overactuated} in
robotics) for all $m\ge3$ to which the pseudo inverse method is
applicable. Moreover, it is easy to verify that the rank of Jacobian is
$\le2$ for all $m>3$ due to the same reasons for (\ref{eq:jacob}), even
when the DPC has many stages. It implies immediately that the matrix
$JJ^T$ is noninvertible but the MP inverse of $J$ can only be evaluated
via, e.g. the singular value decomposition (SVD)\footnote{There are
methods to address the singular $JJ^T$ explicitly, such as the
regularized least squares $\Delta\boldsymbol{\varphi} = J^T(JJ^T +
\lambda I)^{-1} \Delta S_\text{out}$ which minimizes the sum
$\|J\Delta\boldsymbol{\varphi} -\Delta S_\text{out}\|^2 +
\|\Delta\boldsymbol{\varphi}\|^2$.}. The manipulability index $m=
\sqrt{\operatorname{det} (JJ^T)}$ is also zero. The fact we have a
Jacobian with only $2$ degrees-of-freedom regardless of the number of
DPC stages is in sharp contrast to the classical kinematics problem.


\subsection{Gradient projection method} The underdetermined $m$-stage
DPC modeled by (\ref{eq:ctrl_mod_lin}) means there is no unique solution
for $\Delta\boldsymbol{\varphi}$. But all feasible solutions differ by a
vector in the null space of $J$, i.e., $\Delta\boldsymbol{\varphi} =
\Delta \boldsymbol{\varphi}_0 - N\delta$, where
$\Delta\boldsymbol{\varphi}_0$ is an initial solution, the columns of
$N$ are the null-space basis vectors of $J$, and $\delta$ is an
arbitrary vector. Therefore, It is possible to impose further
constraints on $\boldsymbol{\varphi}$ when searching for the optimal
$\Delta\boldsymbol{\varphi}$ by exploiting this degree of freedom in the
null space. The \emph{gradient projection method} chooses a particular
$N\delta$ as the projection of $\boldsymbol{\varphi}$ onto the null
space of $J$ to minimize the norm of $\boldsymbol{\varphi}$. Namely,
\begin{align}
\Delta\boldsymbol{\varphi} 
&= J^{+}\Delta S_\text{out} - NN^{+}\boldsymbol{\varphi}\\ 
&= J^{+}\Delta S_\text{out} - (I-J^{+}J)\boldsymbol{\varphi}\,,
\label{eq:grad_proj}
\end{align}
where $\Delta S_\text{out}$ is again defined by (\ref{eq:ctrl_mod_err}).
Note that there is no need to actually compute the matrix $N$ by using
(\ref{eq:grad_proj}), which constitutes the initial solution $\Delta
\boldsymbol{\varphi}_0 = J^{+} \Delta S_\text{out}$ and the null-space
term $\Delta\boldsymbol{\varphi}_1 = (I - J^{+}J) \boldsymbol{\varphi}$.
Since the linear model is assumed for a local space, which may not
contain the actual minimum $\boldsymbol{\varphi}$, one may take a small
step towards $\Delta\boldsymbol{\varphi}_1$ such that
$\Delta\boldsymbol{\varphi} = \Delta \boldsymbol{\varphi}_0 +\mu
\Delta\boldsymbol{\varphi}_1$ converges to the globally optimal
$\boldsymbol{\varphi}$ with $0< \mu <1$. The classical way to reset the
control signals without changing the polarization state \cite{noe1988}
is similar to null-space operation discussed here. However, it is a
complicated task to find such reset strategies when working with the
nonlinear mapping given by (\ref{eq:ctrl_mod}).


\subsection{Extended Jacobian method} The gradient projection method can
be viewed as the solution to minimize the cost function $H =
\|\boldsymbol{\varphi}\|^2$ subject to the equality constraint given by
(\ref{eq:ctrl_mod}). At the point of $H$ reaching its minimum for the
underdetermined system, $H$ should not change over small variations in
the null space of system Jacobian. Namely, the projection of the
gradient of $H$ onto the null space is zero. Therefore, at those points,
the system satisfies the following equations \cite{chang1987}, with dot
operation for inner product,
\begin{equation}
\label{eq:ext_jacob}
\begin{bmatrix}
f(\boldsymbol{\varphi})\\ N(\boldsymbol{\varphi}) \cdot
\frac{\partial H}{\partial \boldsymbol{\varphi}}
\end{bmatrix} = \begin{bmatrix}
S_\text{out}\\ \mathbf{0}
\end{bmatrix}\,,
\end{equation}
which can be linearized at $\boldsymbol{\varphi}_o$ as
\begin{equation}
\label{eq:ext_lin}
\underbrace{\begin{bmatrix}
J\\ \frac{\partial}{\partial \boldsymbol{\varphi}}
\left[N(\boldsymbol{\varphi}) \cdot
\frac{\partial H}{\partial \boldsymbol{\varphi}}\right]
\end{bmatrix}}_{J_e} \Delta \boldsymbol{\varphi} = 
\begin{bmatrix}
\Delta S_\text{out}\\ -N(\boldsymbol{\varphi}_o) \cdot
\frac{\partial H}{\partial \boldsymbol{\varphi}_o}
\end{bmatrix}\,.
\end{equation}
We obtain a square matrix, called the extended Jacobian $J_e$, on the
left hand side when $J$ is of full rank. The solution of these equations
are found by taking the direct inversion of $J_e$ if it is nonsingular.

\begin{figure}[t]
\centering
\includegraphics{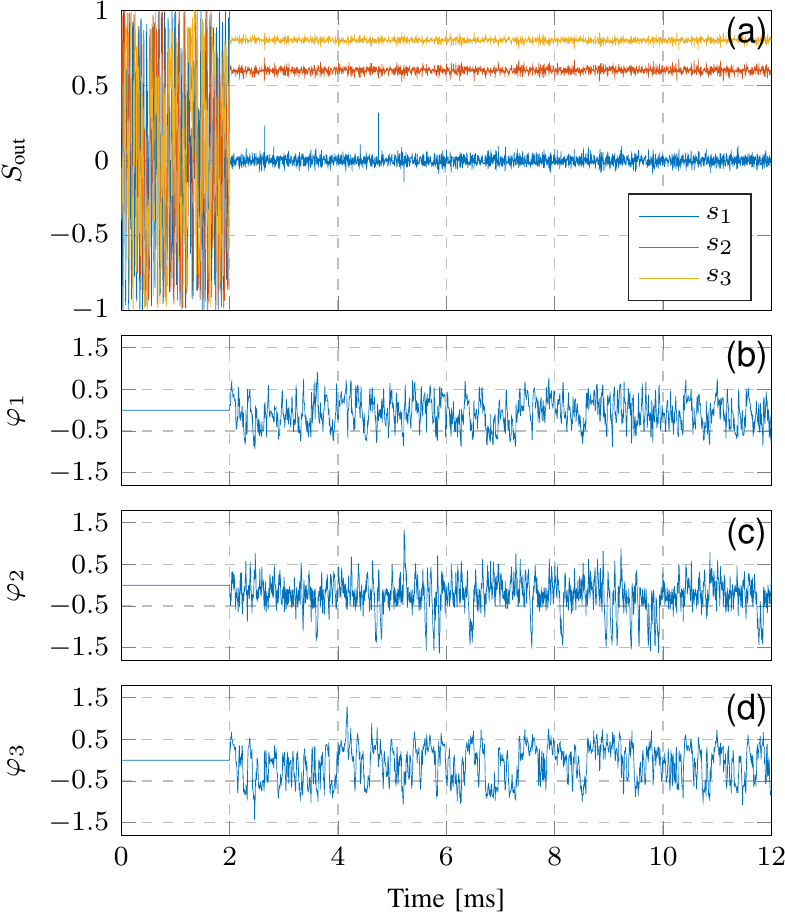}
\caption{Simulation results of using a $3$-stage DPC for locking the
scrambled input SOP to the designated output SOP at $[0,0.6,0.8]^T$. (a)
The time evolution of the Stokes parameters at the output (DPC activated
after $2$ ms). (b-d) The control signal for each stage.}
\label{f3}
\end{figure}

Note that ${\partial H}/ {\partial \boldsymbol{\varphi}}$ is simply the
$\boldsymbol{\varphi}$ vector for $H= \|\boldsymbol{\varphi}\|^2$, which
means the desired $\boldsymbol{\varphi}$ should be orthogonal to the
null space of $J$ according to the formulation in (\ref{eq:ext_jacob}).
When we ignore the second-order term $\partial N(\boldsymbol{\varphi})
/\partial \boldsymbol{\varphi}$, (\ref{eq:ext_lin}) reduces to
\begin{equation}
\label{eq:ext_lin_simp}
\begin{bmatrix} J\\ N^T \end{bmatrix}
\Delta \boldsymbol{\varphi} = 
\begin{bmatrix}
\Delta S_\text{out}\\ -N(\boldsymbol{\varphi}_o) \cdot
\boldsymbol{\varphi}_o
\end{bmatrix}\,.
\end{equation}
A simple method to compute the null-space basis vector $N$ for a
$4$-stage DPC is given by \cite{chang1987}
\begin{equation}
\begin{split}
N &= \begin{bmatrix}
\Delta_1,\Delta_2,\Delta_3,\Delta_4
\end{bmatrix}\\
\Delta_i = (-1)^{i+1} \operatorname{det} 
&(J^1,J^2,...,J^{i-1},J^{i+1},...,J^4)
\,,
\end{split}
\end{equation}
where $\operatorname{det}(\cdot)$ is the determinant with $J^i$ the
$i$th column of $J$. However, the standard extended Jacobian cannot be
used for the DPC model when $J$ is not full rank, which leads to a
singular $J_e$ in (\ref{eq:ext_lin}) and (\ref{eq:ext_lin_simp}).

\subsection{Special cases} The analysis we have given so far is fully
general whereas special cases exist in real applications. For instances,
the desired SOP target may not be a single point on the Poincar\'{e}
sphere but a particular region, say, all points with $s_3 = 0$. In such
cases, the linear DPC model has an output space with one dimension and
an $1 \times m$ Jacobian. The gradient projection method is applicable
and, since $J$ now has full rank, we should be able to find $m-1$ basis
vectors in its null space and get an invertible extended Jacobian.

\section{Results and discussion} We conduct numerical simulations by
sending a polarization scrambled light into the DPC with Euler-angles
composition. The input SOP is characterized by the rotation about the
axis $S_3$ with a constant rate of $100$ krad/s, perturbed by the
rotation about the axis $S_2$ with randomly distributed angels. The
acquisition rate of $S_\text{out}$ is set to be $50$ MHz and the DPC is
activated after $2$ ms. The overall processing delay $\tau$ of the
closed-loop DPC is assumed to be $10^{-6}$ second, meaning that the
$\Delta \boldsymbol{\varphi}$ obtained at time $t$ is used to update the
control signal at $t + \tau$.

\begin{figure}[t]
\centering
\includegraphics{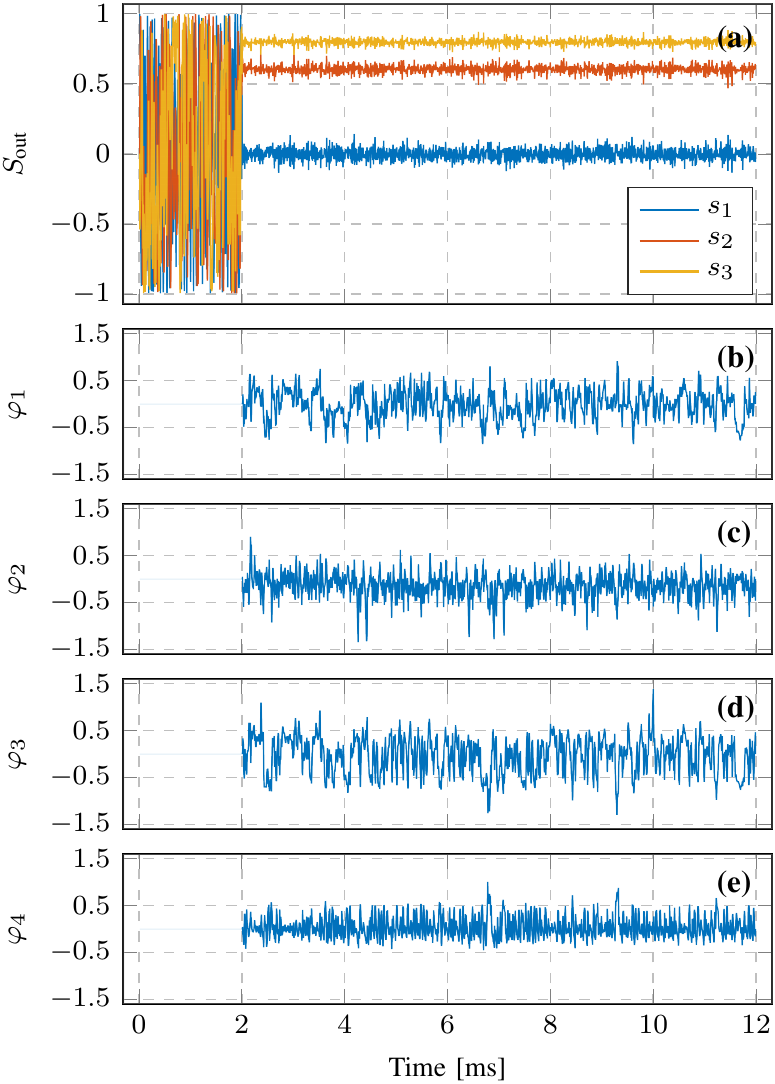}
\caption{Simulation results of using a $4$-stage DPC for locking the
scrambled input SOP to the designated output SOP at $[0,0.6,0.8]^T$. (a)
The time evolution of the Stokes parameters at the output (DPC activated
after $2$ ms). (b-e) The control signal for each stage.}
\label{f4}
\end{figure}

First, we study a $3$-stage DPC system with the composition given by $M
= R_1{R_3}R_1$ and the $3$ control signals are linearly related to the
rotation angles via $\theta=\pi\cdot\varphi$, i.e., one unit of control
signal generates $\pi$ phase shift. The target SOP is $[0,0.6,0.8]^T$.
The gradient projection method is used to obtain the phase variations
with a step size of $0.1$. The time evolution of the individual entries
of the vector $S_\text{out}$ is shown in Fig. \ref{f3}(a), showing
successful SOP convergence is rapidly achieved when the DPC is activated
after $2$ ms. The required control signals, as shown in Fig. 3(c), are
clearly well bounded (mostly $\in (-1.5,1.5)$). The target SOP is chosen
for better illustration only. However, we do observe that the control
signals have more outliers with large values when the $S^*_\text{out}$
coincides with the rotation axis of the last stage, which zeros the last
column of Jacobian such that it is more likely to have a small second
singular value after the SVD. The outliers also represent the situations
when the DPC is close to loss another DOF and become a $1$-rank system.

The control performance can be improved by using a $4$-stage DPC. The
results based on the composition $M = R_1R_3R_1R_3$  and the same
settings as above are shown in Fig. \ref{f4}. We obtain now a better SOP
convergence and control phases varying over a slightly smaller range.

\begin{figure}[t]
\centering
\includegraphics{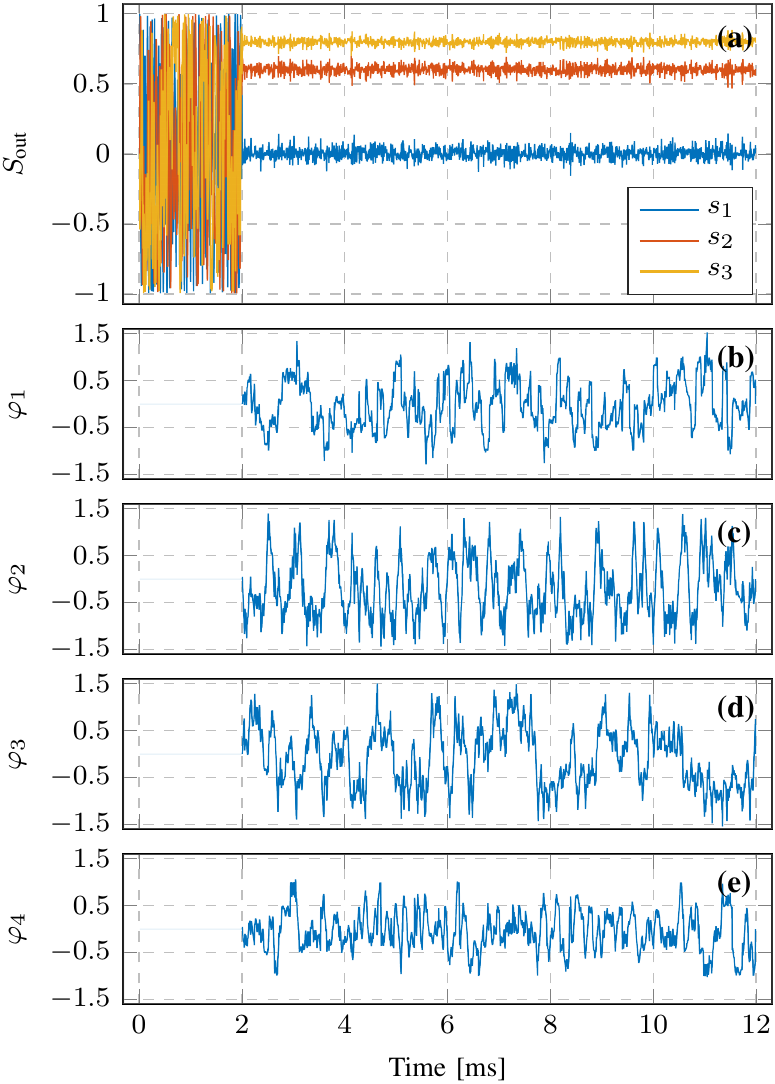}
\caption{Simulation results of using a $4$-stage DPC for locking the
scrambled input SOP to the designated output SOP at $[0,0.6,0.8]^T$. The
gradient projection is only included when the controls signals exceed
a threshold. (a-e) The same plotting as that in Fig. \ref{f4}.}
\label{f5}
\end{figure}

The computational complexity of the gradient projection method lies in
the computation of Jacobian matrix, the SVD-based pseudo inverse and the
null space correction term. In the examples we present here, the
Jacobian has a relatively simple form but does need real-time update
since it depends on the varying input SOP explicitly. The pseudo inverse
seems inevitable and requires efficient algorithm to implement. On the
other hand, we need not to include the null space term all the time
unless the control signals are close to their boundaries. We study the
case by only activating the null-space correction when one of the
control signals exceeding $\pm 1$. The results are shown in Fig.
\ref{f5}. To compare with the results in Fig. \ref{f4}, it turns out the
null-space correction needs only to be included about $20\%$ of the
operation time without sacrificing the DPC performance. The percentage
clearly depends on the SOP varying speed at the DPC input. We note that
the gradient projection can be activated much like the classical signal
resetting, but in a more systematic way.

The Euler-angles composition is obviously just one of many possible DPC
realizations. For other DPC solutions, it might be difficult, if not
impossible, to get a close-form Jacobian so that techniques like
dithering becomes essential. As results, the Jacobian condition may not
be as clear as the examples shown in this article. However, the gradient
projection method applies to any type of DPC as long as it can be
modeled by an underdetermined linear model. The extended Jacobian
method, on the other hand, asks for a Jacobian matrix with full rank.

\section{Conclusion} We have discussed several Jacobian-based algorithms
for fast and reset-free dynamic polarization control (DPC). By
linearizing the general mapping from the control space to the Stokes
space, we have shown that the DPC problem can be linked closely to the
arm-type robot kinematics. Methods including the gradient projection and
the extended Jacobian are analyzed based on a popular DPC realization.
We have analyzed in detail on the Jacobian condition of the DPC modeled
as multiple series-lined waveplates with Euler-angles composition. We
confirm the Jacobian has rank deficiency when the number of waveplates
is larger than or equal to $3$. Numerical simulations of gradient
projection demonstrate the usage of null-space correction for confining
the physical range of control signal. The results of this work can be
generalized for other DPC models, the Jacobian condition of which
however demands a case-by-case analysis.

\ifCLASSOPTIONcaptionsoff
	\newpage
\fi

\bibliographystyle{IEEEtran}
\bibliography{IEEEabrv,mybib}




\end{document}